\documentclass{article}

\usepackage{PRIMEarxiv}
\usepackage{float} 
\usepackage[T1]{fontenc}    
\usepackage{hyperref}       
\usepackage{url}            
\usepackage{booktabs,tabularx,array}       
\usepackage{amsfonts}       
\usepackage{nicefrac}       
\usepackage{microtype}      
\usepackage{lipsum}
\usepackage{fancyhdr}       
\usepackage{graphicx}       
\graphicspath{{media/}}     
\usepackage{amsmath}        
\usepackage{amssymb}        
\usepackage{subcaption}     

\floatstyle{ruled}
\newfloat{algorithm}{htbp}{loa}
\floatname{algorithm}{Algorithm}

\title{TTT in VLA Models: Exploring Training with Human Demonstrations}

\author{
    Qiangong Zhou\thanks{These authors contributed equally to this work.}
    \quad
    Jiafeng Liang\footnotemark[1]
    \quad
    Yimin Dong\footnotemark[1]
    \quad
    Yang He
    \quad
    Hongxiang Li\thanks{Corresponding author:
    \texttt{hongxiang.li@standard-robots.com}.}
    \\[0.5em]
    Standard Robots Co., Ltd.
    \\[0.3em]
    {\small
    \texttt{zhouqiangong@standard-robots.com}
    \quad
    \texttt{liangjiafeng@standard-robots.com}}
    \\
    {\small
    \texttt{dongyimin@standard-robots.com}
    \quad
    \texttt{heyang@standard-robots.com}}
    \\
    {\small
    \texttt{hongxiang.li@standard-robots.com}}
}

\begin{document}
\maketitle

\begin{abstract}
Long-horizon robot tasks require retaining and using past information. We integrate Test-Time Training (TTT), which stores context in fast weights, into InternVLA-A1 and Pi0.5 to study action generation conditioned on human demonstrations through training on paired demonstrations and robot trajectories. We implement memory transfer between the two phases and truncated backpropagation, and examine training objectives, gradient coverage, and contrastive regularization across video, short-video, and single-frame cue tasks.

In our experiments, adding TTT and training on paired data at the current scale did not lead to consistent target selection based on demonstrations. Under the tested settings, shorter sequences, full-trajectory gradient connectivity, and contrastive regularization did not clearly improve responsiveness to demonstrations. When demonstrations were reduced to single-frame cues, real-robot execution still showed a preference for cups at fixed positions. We document the implementation, training configurations, and observations to support further inspection and replication.

\end{abstract}


\section{Introduction}
Our overall goal is to explore the memory capabilities needed for long-horizon robot tasks. Later decisions may depend on information encountered much earlier, so models need to retain and use historical context. We therefore consider Test-Time Training (TTT)\cite{1}. TTT writes past information into a fixed-size memory state by updating fast weights. Unlike caches that grow with history length, its memory-state storage remains constant during inference, offering a substantial advantage for long-horizon tasks. TTT also supports online updates and end-to-end learning of memory writes and reads. These properties motivate our exploration.

We focus on robot action generation conditioned on human demonstrations. Each training sample pairs a human demonstration with its corresponding robot trajectory. The model first reads the demonstration, then processes the robot trajectory using the resulting memory. TTT provides the memory mechanism, while task training teaches the model how demonstration conditions should affect action generation. We implement a standard TTT layer and integrate it into Pi0.5\cite{2} and InternVLA-A1\cite{3}. Following the TBPTT approach in RoboTTT\cite{4}, we organize state transfer and gradient propagation between human demonstrations and robot trajectories.

Our main contribution is to reproduce and clearly describe the TTT architecture and its training procedure for combining human demonstrations with robot trajectories in VLA models. Training infrastructure constraints have limited the scale of pretraining and training on memory-task data, leaving TTT's memory capabilities insufficiently trained and evaluated. We also report observations across task settings and discuss training objectives, evaluation protocols, and contrastive regularization. We leave unresolved causes and incomplete validation explicitly open.

\section{Related Work}

\subsection{From Test-Time Adaptation to Parametric Memory}
\label{sec:2.1}

Early Test-Time Training (TTT)\cite{5} primarily addressed distribution shifts between training and test data. The model constructs a self-supervised task from unlabeled test inputs and updates its parameters before prediction. TTT++\cite{6} and TTT-MAE\cite{7} later explored adaptation through feature-distribution alignment and masked image reconstruction, respectively.

Subsequent work incorporated inner learning into the model architecture. MTTT\cite{8} learns the inner self-supervised task through outer-loop task training. TTT layers\cite{1} use the inner model's weights as recurrent state, continually compressing historical context into those weights. Titans\cite{9} and TTT-E2E\cite{10} also explore retaining context through test-time parameter updates, treating these updates as memory writes.

In the TTT layer\cite{1} used here, the model constructs \(K\), \(V\), and \(Q\) from the input. The inner model writes information into fast weights by learning to predict \(V\) from \(K\), then reads it using \(Q\). Outer-loop task training learns how this information should be written and used. The inner model, objective, and update rule jointly define the design space.

\subsection{TTT across Domains}
\label{sec:2.2}

In language modeling, TTT-Linear and TTT-MLP\cite{1} store historical state in a linear model and a two-layer MLP, respectively. Titans\cite{9} combines long-term neural memory with local attention. TTT-E2E\cite{10} updates weights through next-token prediction and learns their initialization through meta-training. These methods all absorb context through parameter updates, but differ in their memory structures and inner objectives.

In vision, TTT-MAE\cite{7} uses masked reconstruction for test-time adaptation, while MTTT\cite{8} uses inner learning as a visual representation module. ViT³\cite{11} further examines inner models, losses, and update rules, including full-batch updates, inner learning rates, model capacity, and convolutional structures. These choices inform the implementation of visual TTT.

For video, TTT on Video Streams\cite{12} carries forward previously updated parameters and adapts online to incoming frames. One-Minute Video Generation\cite{13} retains local attention in a pretrained video model and uses TTT to transfer context across segments. These methods illustrate different ways to process temporal information, but their training procedures still require adaptation to robot tasks.

\subsection{TTT for Robot Policies}
\label{sec:2.3}

RoboTTT\cite{4} introduces fast-weight memory into an action model and combines Sequence Action Forcing with TBPTT for sequence training. Human videos update memory without providing action supervision; subsequent robot trajectories inherit that memory and contribute action losses. WAM-TTT\cite{14} instead adds TTT to the video branch, updating memory through video prediction and key–value reconstruction and learning its connection to actions through task losses on paired robot data.

Recent work further explores how memory is organized. RecastVLA\cite{15} updates shared fast weights using action-generation features. T²Mem\cite{16} forms memory from observations and learns to retain and use it through action supervision. TTT-RM\cite{17} uses historical reconstruction residuals so that fast weights can supplement information missing from current observations and explicit memory. These studies examine memory interfaces, update timing, and learning objectives.

Drawing on these memory mechanisms and training approaches, we integrate TTT into Pi0.5\cite{2} and InternVLA-A1\cite{3} and explore training with human demonstrations and robot trajectories. We focus on how demonstration information enters fast weights and how robot supervision constrains its use. The following sections describe the implementation and experimental observations.

\section{Architecture}
\label{sec:3}

This section describes the TTT layer, its integration into VLA backbones, and the corresponding TBPTT training procedure. We first outline sequence training in RoboTTT, then describe the connection between human demonstrations and robot trajectories and the independent configuration of their memory-write frequencies.

\subsection{TTT Layer Structure}
\label{sec:3.1}

\textbf{Fast and slow weights.} All parameters of the original VLA network are slow weights in our architecture. They are updated during outer-loop task training and remain fixed during inference. The added TTT layer contains both slow and fast weights. Its projection, normalization, and gating parameters are slow weights, while the runtime parameters of its inner two-layer MLP are fast weights.

Fast weights carry memory. As inputs are processed, inner learning updates these weights to compress context into them. Outer-loop task training teaches the slow weights what to write and how to read and use it. The two types of weights thus serve to store current context and learn the memory mechanism, respectively. Here, "fast" and "slow" distinguish update mechanisms rather than learning-rate magnitudes.

\begin{figure}[htbp]
\centering
\includegraphics[width=0.6\linewidth,height=0.58\textheight,keepaspectratio]{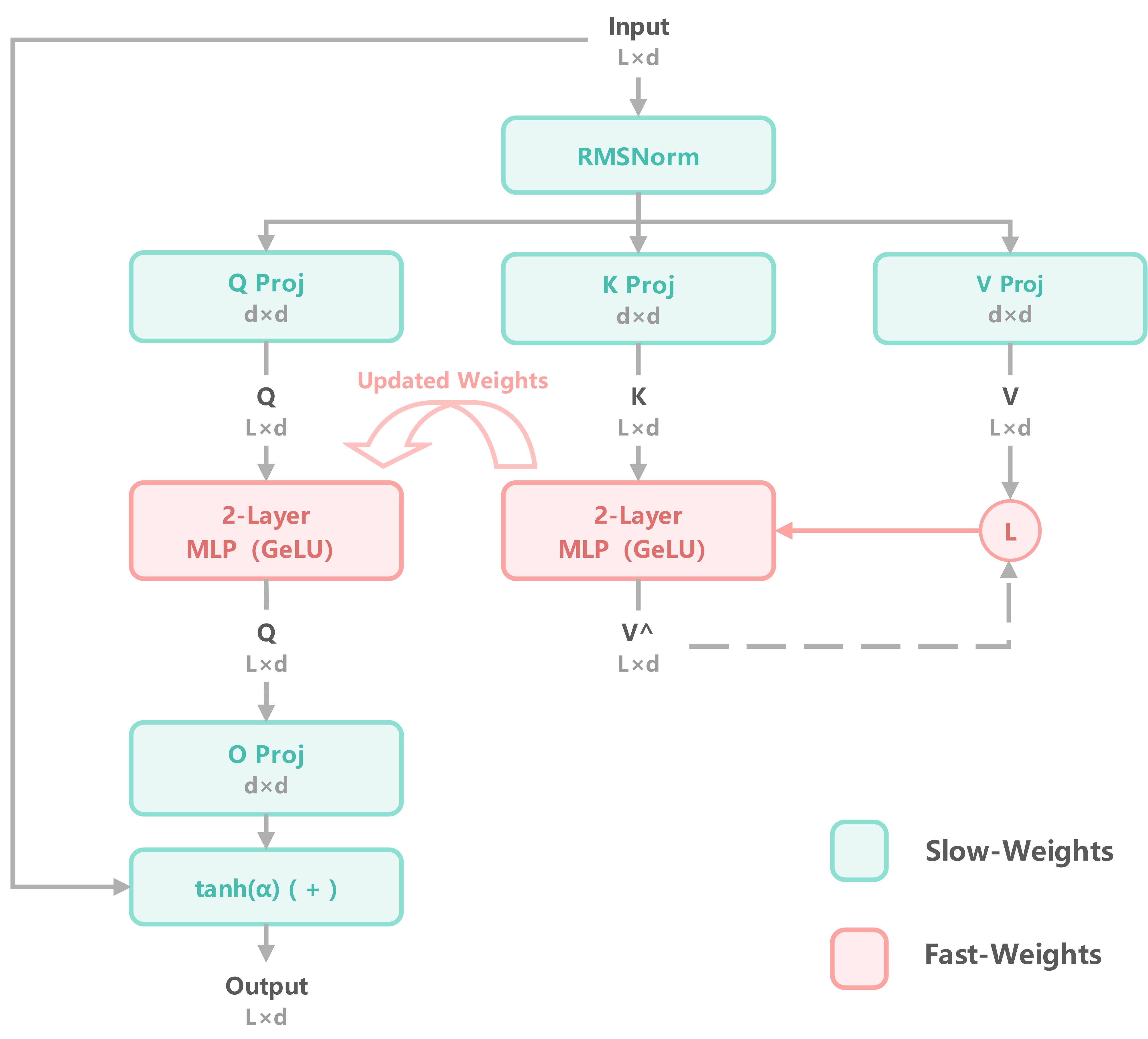}
\caption{\textbf{TTT layer architecture and memory operations.} Green modules contain slow weights, while red modules represent a two-layer GELU MLP parameterized by fast weights. Normalization and projections transform the input into \(Q\), \(K\), and \(V\). The right path predicts \(V\) from \(K\) and updates fast weights using the inner loss. The left path queries the updated model with \(Q\), then applies an output projection and a gated residual connection. The two MLPs depict two calls to the same inner model, before and after the update. The batch dimension, head splitting, and inner normalization and residual details are omitted for clarity.}
\label{fig:ttt-layer}
\end{figure}


\textbf{Write and query representations.} As shown at the top of Figure~\ref{fig:ttt-layer}, the input \(X\in\mathbb{R}^{L\times d}\) passes through RMSNorm and three independent projections:

\begin{equation}
\label{eq:qkv}
\bar X=\operatorname{RMSNorm}(X), \qquad Q=\bar XW_Q,\quad K=\bar XW_K,\quad V=\bar XW_V.
\end{equation}

Here, \(L\) is the number of input tokens and \(d\) is the feature dimension. We omit the batch dimension and head splitting for simplicity. \(W_Q\), \(W_K\), and \(W_V\) are TTT slow weights, independent of the surrounding attention module's projection parameters.

The three representations have distinct roles: \(K\) is the inner learning input, \(V\) is its target, and \(Q\) queries the resulting memory. All three come from the current input. In particular, \(V\) is produced by a learnable projection and requires no additional external labels.

\textbf{Writing memory through inner learning.} The right side of Figure~\ref{fig:ttt-layer} shows the write operation. We use a two-layer GELU MLP as the inner model, denoted by \(g_F\), with parameters \(F\). Each head maintains its own \(F\) and updates it across subsequent calls.

Following Section 2.7 of the original TTT paper\cite{1}, the inner model uses a reconstruction objective with normalization and a residual connection:

\begin{equation}
\label{eq:inner-objective}
\widehat V = K+\operatorname{Norm}\bigl(g_F(K)\bigr), \qquad \mathcal L_{\mathrm{inner}} = \operatorname{MSE}(\widehat V,V).
\end{equation}

The inner MLP learns the correction needed to transform \(K\) into \(V\). The prediction error determines how the current input changes the fast weights. We then take one gradient step:

\begin{equation}
\label{eq:fast-update}
F^{+}
=
F-\eta\nabla_F\mathcal L_{\mathrm{inner}}.
\end{equation}

This step corresponds to the red update arrow in Figure~\ref{fig:ttt-layer}. The current implementation fixes the inner learning rate at \(\eta=1\) and jointly computes the inner loss over the tokens selected for writing in the current call. The updated \(F^{+}\) becomes the new memory state.

\textbf{Reading memory and producing the output.} The left side of Figure~\ref{fig:ttt-layer} shows the read operation. After the write, the model feeds \(Q\) into the updated inner model:

\begin{equation}
\label{eq:memory-read}
U
=
Q+\operatorname{Norm}\bigl(g_{F^{+}}(Q)\bigr).
\end{equation}
The readout depends on both the current query \(Q\) and the fast weights \(F^{+}\). Changes to fast weights caused by earlier inputs can thus affect the current output.

The head outputs are merged, projected, and added back to the input through a gated residual connection:

\begin{equation}
\label{eq:gated-output}
Y
=
X+\tanh(\alpha)\odot\operatorname{OProj}(U).
\end{equation}

The output projection and gate parameters \(\alpha\) are slow weights. The projection maps the readout back to the model's feature space, while the gate controls the contribution of the TTT branch. The inner \(Q\) residual participates in memory readout, whereas the outer \(X\) residual connects the entire TTT branch to the original model.

\textbf{Learning the memory mechanism through slow weights.} The inner loss updates runtime fast weights, while the outer task loss trains slow weights. During outer-loop training, gradients pass through memory reads and differentiable fast-weight updates to train the Q/K/V projections, output projection, and other unfrozen parameters. Task supervision can therefore teach the model what to write, how to query memory, and how to use its readout.

The initial fast weights \(F_{\mathrm{init}}\) are also learned through outer-loop training. Each new trajectory starts from this initialization and develops runtime fast weights as inputs arrive. The initialization is a learnable parameter, while the runtime weights are the memory state of the current trajectory.

Figure~\ref{fig:ttt-layer} shows the “update and read” mode: the model first updates fast weights using the current input, then reads the updated memory. In “read-only” mode, fast weights remain unchanged and the model only queries memory and computes the output. State transfer across anchors and gradient truncation during training are described in Section~\ref{sec:3.3}.

\begin{algorithm}[htbp]
\caption{TTT Layer: Memory Update and Readout}
\label{alg:ttt-layer}
\small
\noindent\textbf{Require:} Input \(X\), fast weights \(F\), write-token set \(W\), mode \(m\)\par
\noindent\textbf{Require:} Inner learning rate \(\eta\), number of heads \(H\)\par
\noindent\textbf{Ensure:} Output \(Y\), updated fast weights \(F'\)\par
\begin{enumerate}
\renewcommand{\labelenumi}{\arabic{enumi}:}
\setlength{\itemsep}{1pt}
\setlength{\parskip}{0pt}
\setlength{\parsep}{0pt}
\item \textbf{Step 1: Input projection}
\item \(\bar X \gets \operatorname{RMSNorm}(X)\)
\item \(Q,K,V \gets \operatorname{SplitHeads}(\bar X W_Q,\bar X W_K,\bar X W_V)\)
\item \textbf{Step 2: Memory update}
\item \textbf{for} \(h=1,\ldots,H\) \textbf{do}
\item \hspace*{1em}\(F'_h \gets F_h\)
\item \hspace*{1em}\textbf{if} \(m=\text{\textsc{UpdateAndRead}}\) and \(W\ne\varnothing\) \textbf{then}
\item \hspace*{2em}\(K_w,V_w \gets K_h[W],V_h[W]\)
\item \hspace*{2em}\(\mathcal L_{\mathrm{inner}} \gets \operatorname{MSE}(K_w+\operatorname{Norm}(\operatorname{MLP}(K_w;F_h)),V_w)\)
\item \hspace*{2em}\(F'_h \gets F_h-\eta\nabla_{F_h}\mathcal L_{\mathrm{inner}}\)
\item \hspace*{1em}\textbf{end if}
\item \textbf{end for}
\item \textbf{Step 3: Memory readout}
\item \textbf{for} \(h=1,\ldots,H\) \textbf{do}
\item \hspace*{1em}\(U_h \gets Q_h+\operatorname{Norm}(\operatorname{MLP}(Q_h;F'_h))\)
\item \textbf{end for}
\item \textbf{Step 4: Gated residual output}
\item \(U \gets \operatorname{MergeHeads}(U_1,\ldots,U_H)\)
\item \(Y \gets X+\tanh(\alpha)\odot\operatorname{OProj}(U)\)
\item \textbf{return} \(Y,F'\)
\end{enumerate}
\end{algorithm}



The fast-weight update in Line 10 of Algorithm~\ref{alg:ttt-layer} remains differentiable during outer-loop training, allowing task losses to propagate back to memory writes. Each sample and head maintains independent fast weights. The algorithm follows Section~\ref{sec:3.1}; the constrained Fast MLP variant in Pi0.5 replaces the inner objective and update rule as described in Section~\ref{sec:3.2.2}.

\subsection{TTT Layer Integration}
\label{sec:3.2}

We next describe how TTT is integrated into the two backbone models.

\subsubsection{InternVLA-A1}
\label{sec:3.2.1}

\textbf{Backbone and insertion point.} InternVLA-A1\cite{3} has understanding, generation, and action experts for semantic understanding, visual prediction, and action generation, respectively. Each expert has separate attention projections, an output projection, and a feedforward network. At each layer, the experts independently produce queries, keys, and values, then concatenate them along the sequence dimension for joint attention. The attention output is split by expert and returned to the corresponding network.

We insert a TTT layer into every layer of both the generation and action experts, leaving the understanding expert unchanged. Following RoboTTT\cite{4}, TTT sits after the attention sublayer and before the original feedforward sublayer. Specifically, the generation and action experts apply their output projections to the joint-attention results and complete the first residual connection. The resulting features enter the TTT layer in Figure~\ref{fig:ttt-layer} for memory updates and reads, then pass through the original normalization and feedforward network before the second residual connection.

TTT therefore receives features already fused by attention in the current layer. It constructs its own write and query representations and adds the readout to subsequent computation through a gated residual connection. This residual connection is internal to the TTT layer, whose output is used directly by the surrounding network.

\textbf{Expert selection.} The generation expert predicts future visual information, while the action expert generates robot actions. Adding memory only to the action expert would leave the two branches conditioned on different histories. We therefore add TTT to both experts so that each can write and use memory. This design is intended to let historical context influence both visual prediction and action generation; its effects still require experimental evaluation. All stages of our InternVLA-A1 experiments use this integration in every layer.

Each enabled insertion point maintains independent fast weights. Within a layer, weights are also separate across heads and samples. Each trajectory updates its state chronologically and shares no runtime memory with other trajectories. Each forward pass receives existing fast weights and returns an updated state for subsequent anchors. Training and inference use the same TTT insertion points.

\textbf{Write inputs in the action expert.} The action expert receives robot-state tokens and noisy action tokens. We considered three write configurations and use the third in this report. The first allows both token types to write to fast weights, so updates receive both state and action representations. The second permits only state tokens to write; action tokens read memory but do not contribute to the inner reconstruction objective.

The third configuration adds learnable register tokens between the state and action tokens. The registers aggregate current input information through attention and write memory together with state tokens; action tokens only read memory. To make this route explicit, we restrict direct attention from action tokens to registers, so register information affects action representations through the TTT branch.

\textbf{Memory operations during denoising.} Register and state tokens write memory, while action tokens only read it. Each observation anchor performs one fast-weight update; all subsequent denoising iterations for action generation read the same memory without further writes. The write frequency is therefore determined by observation anchors and is independent of the number of denoising iterations.

After the current action chunk, fast weights pass to the next observation anchor. A new trajectory resets them to the learnable initialization. Fast weights and attention key–value caches are managed separately: the former carry memory across anchors, while the latter reuse attention representations. Memory transfer between human demonstrations and robot trajectories is described in Section~\ref{sec:3.3}.

\subsubsection{Pi0.5}
\label{sec:3.2.2}

\textbf{Backbone and insertion point.} We build the memory policy on the OpenPI implementation of Pi0.5\cite{2}. Image and text conditions form prefix representations, and the action expert generates action chunks through flow matching. TTT is inserted into all 18 action-expert layers, after the attention output projection and residual connection and before the original feedforward network. Historical readouts enter the current representation through a gated residual connection, directly involving memory in action generation. The visual encoder and prefix language model retain their original structures, and no additional visual generation expert is introduced.

\textbf{Write inputs and attention paths.} Each observation anchor uses a separate write branch containing 16 learnable register tokens and one continuous robot-state token. These tokens aggregate the current prefix information through attention and jointly write to fast weights. The original state-conditioning mechanism of Pi0.5 is retained. Action tokens read memory but cannot directly attend to the register cache, routing register information through TTT. The action branch can still access the current visual prefix.

\textbf{Decoupling writes from denoising.} The write branch uses a fixed time condition, receives no noisy actions, and does not depend on the sampled denoising timestep \(\tau\). During training, noise and \(\tau\) are sampled for each action chunk to compute the flow-matching loss, whose gradients can pass through fast-weight updates. During inference, each observation anchor writes once, and the following 10 denoising steps all read the same fast weights. This avoids repeated writes of the same observation and keeps write frequencies consistent between training and inference.

\textbf{Constrained Fast MLP.} In Pi0.5, we compare a constrained Fast MLP with the multi-head core used in InternVLA-A1. Both share the insertion points and memory operations described above and connect back to action representations through an output projection and a channel-wise \(\tanh\) gate.

The constrained Fast MLP uses a two-layer \(256\to1024\to256\) MLP at each layer, fitting \(V\) with \(g_W(K)\). Its inner learning rate combines a base coefficient of 0.03 with a learnable coefficient and is capped at 0.05. The norm of the actual update is constrained independently for each trajectory and layer, using a shared scaling factor for parameters within a layer. Thresholds are calibrated per layer and then fixed. Clipping preserves gradients, and training and inference use the same write rule.

The multi-head core used in InternVLA-A1 follows the normalized residual reconstruction in Section~\ref{sec:3.1}. Each layer has 16 heads, each with a two-layer \(64\to256\to64\) MLP and a fixed inner learning rate of 1. Both cores store and update fast weights in FP32; their TTT slow parameters use FP32 and BF16, respectively. Their inner learning rates are not directly comparable because their structures and loss normalization differ.

Fast weights are maintained independently for each trajectory and layer, with new trajectories starting from the learnable initialization \(W_0\). Fast weights carry across anchors, and training truncates gradients only at TBPTT segment boundaries, as described in Section~\ref{sec:3.3}.

\subsection{TBPTT with Human Demonstrations}
\label{sec:3.3}

\subsubsection{From RoboTTT to TBPTT}
\label{sec:3.3.1}

RoboTTT\cite{4} organizes action-chunk prediction as sequence training along a trajectory. At each anchor, the model receives the observation and state, predicts the corresponding action chunk, and passes updated fast weights to the next anchor. Noise and denoising timesteps are sampled independently for different chunks. This Sequence Action Forcing procedure lets the model learn the current chunk while receiving memory from earlier anchors.

Later action losses can influence earlier memory writes through fast-weight updates. Training therefore requires differentiable fast-weight updates. Full backpropagation through a trajectory requires retaining its computation graph. Longer trajectories therefore require more training GPU memory.

RoboTTT addresses this with truncated backpropagation through time (TBPTT). A trajectory is divided into consecutive segments, with computation graphs retained and backpropagation performed within each segment. At a segment boundary, fast weights are detached from the preceding graph, but their values carry forward. Later segments can still read previously written memory, but gradients cannot cross that boundary along the fast-weight recurrence.

Two types of grouping should be distinguished. In Section~\ref{sec:3.1}, the tokens participating in one inner update jointly write to fast weights within a single call. Here, a training segment contains multiple chronological anchors and controls the extent of outer-loop backpropagation. The former defines how a write is computed; the latter defines how far gradients propagate through time.

\subsubsection{Segment Backpropagation and Slow-Weight Updates}
\label{sec:3.3.2}

\begin{figure}[htbp]
\centering
\includegraphics[width=0.95\linewidth,height=0.58\textheight,keepaspectratio]{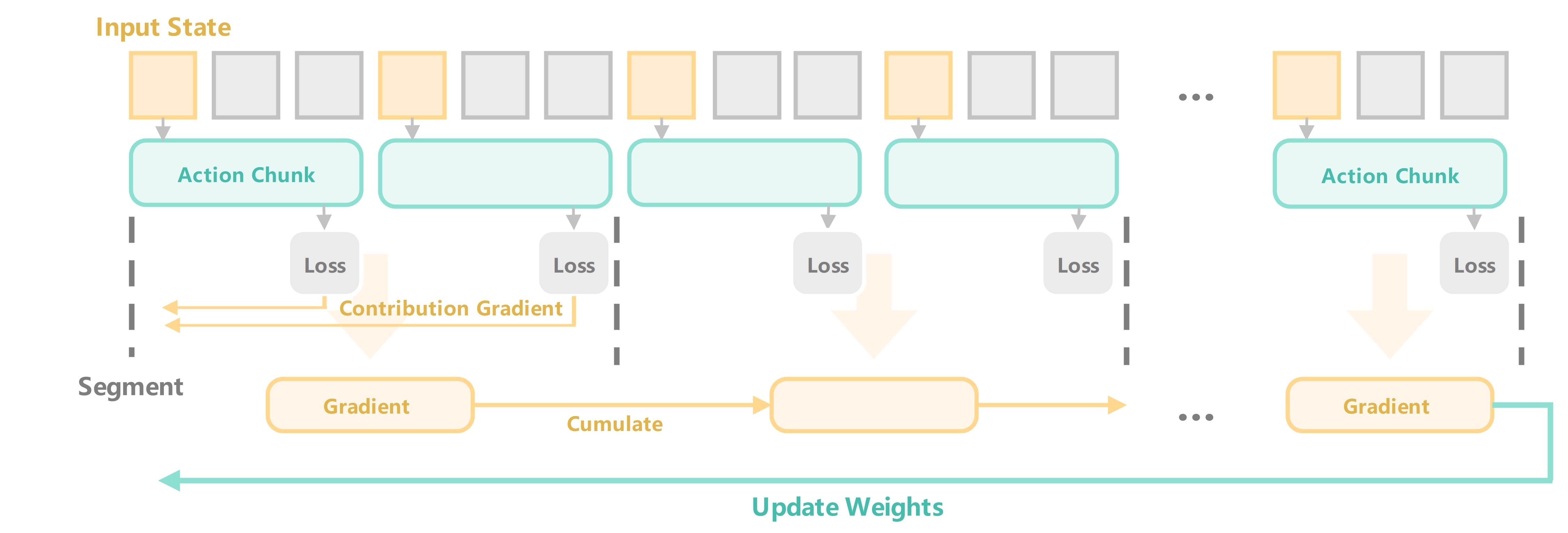}
\caption{\textbf{TBPTT and gradient accumulation in our implementation.} Yellow boxes at the top are sampled decision anchors, each predicting an action chunk and computing its loss. Vertical dashed lines separate segments; yellow arrows indicate gradient contributions within segments and accumulation across segments. After the full trajectory batch is processed, accumulated gradients update slow weights, as indicated by the green arrow at the bottom. Inner fast-weight updates and state transfer across segments are not drawn separately.}
\label{fig:tbptt}
\end{figure}

Our implementation samples robot anchors using consecutive, non-overlapping action chunks. If a chunk contains H control steps, adjacent anchors are H steps apart. Each anchor computes a loss for its action chunk and passes fast weights to later anchors. Consecutive anchors form a training segment, as shown in Figure~\ref{fig:tbptt}.

At the end of each segment, we aggregate its losses, backpropagate, release the segment graph, and detach fast weights from the preceding graph. The fast-weight values remain unchanged and initialize the next segment's memory. Slow-weight gradients accumulate across segments, and one slow-weight update is performed only after the entire trajectory batch is processed.

Within a trajectory batch, slow weights remain fixed while fast weights continue to update across anchors. Accumulating slow-weight gradients across segments does not restore truncated temporal gradient paths. This optimizer update schedule is our training choice, not a requirement of TBPTT itself.

Each trajectory in a batch maintains independent fast weights. When a shorter trajectory ends, its memory is frozen and it no longer contributes writes or losses while the remaining trajectories finish. At the next batch, every trajectory resets its fast weights to the learnable initialization.

\subsubsection{Memory Read and Write Frequencies}
\label{sec:3.3.3}

\begin{figure}[htbp]
\centering
\includegraphics[width=0.85\linewidth,height=0.58\textheight,keepaspectratio]{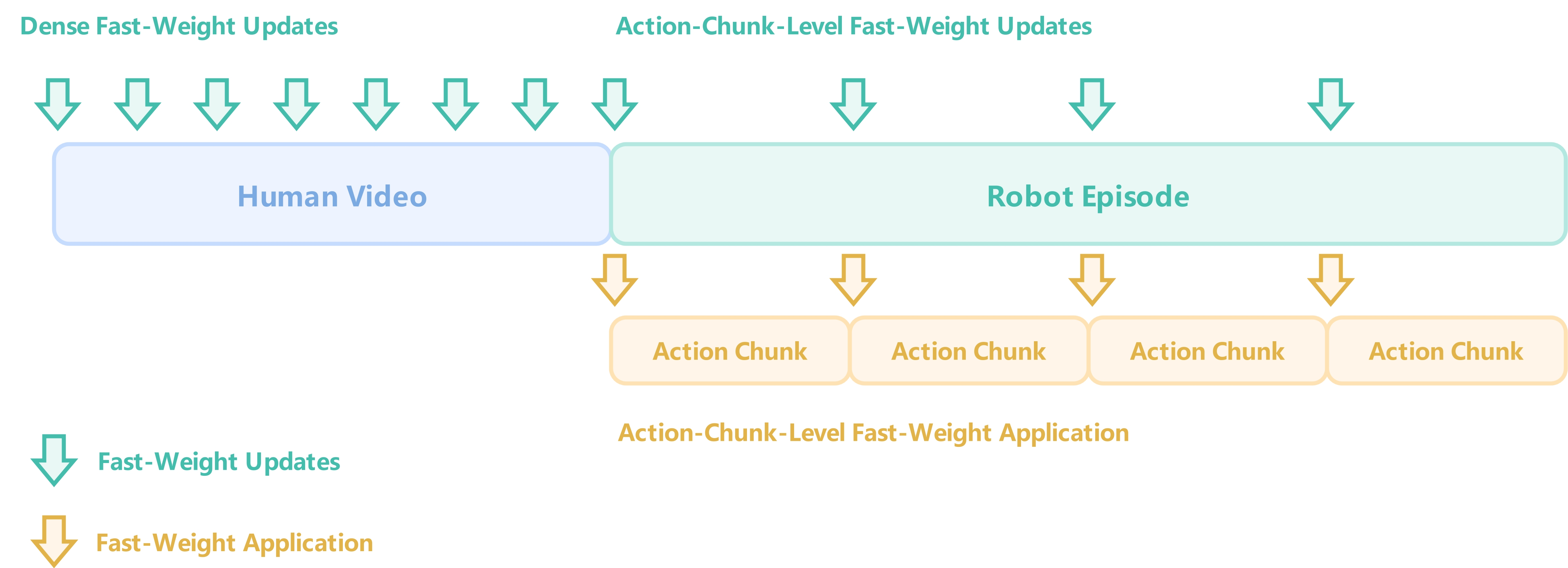}
\caption{\textbf{Memory operations in human demonstrations and robot trajectories.} Blue denotes the human phase, and green denotes the robot phase. Green arrows indicate fast-weight updates; yellow arrows indicate memory use in action-chunk prediction. The human phase may write more frequently, while the robot phase advances at action-chunk intervals. Fast weights carry across the phase boundary. Arrow density only illustrates that the frequencies can differ; it does not specify a fixed ratio.}
\label{fig:human-robot-memory}
\end{figure}

For paired data, we process the human demonstration chronologically before the corresponding robot trajectory. Fast weights carry across the two phases without reinitialization. The demonstration serves as a training condition, and its memory participates in subsequent robot action generation. Figure~\ref{fig:human-robot-memory} illustrates the memory operations in both phases.

During the human phase, the model reads video observations at the configured frequency and updates fast weights through enabled TTT branches. It generates no robot actions and computes no robot action supervision loss, but still writes memory through the inner reconstruction objective. Thus, the absence of yellow action-prediction arrows in Figure~\ref{fig:human-robot-memory} does not imply an absence of forward computation or inner updates.

During the robot phase, the model inherits demonstration memory and predicts action chunks from the current observation and state. Memory continues to update along the robot trajectory, which supplies the training losses. Denoising iterations follow the memory operations described in Section~\ref{sec:3.2}. An action-chunk prediction in Figure~\ref{fig:human-robot-memory} does not imply a memory write at every denoising iteration.

The two phases can use independently configured write frequencies. Human writes are scheduled by video time, while robot anchors advance at action-chunk intervals. With non-overlapping chunks and one write per anchor, the robot write frequency is determined jointly by the control frequency and chunk length. Human demonstrations may use denser writes, but this is not required by the training procedure.

This decoupling also affects the time span of a training segment. The same number of anchors can cover different durations in the two phases. Experimental configurations should therefore report write frequencies and anchor counts separately for each phase; segment length alone does not specify the temporal coverage of memory updates.

\subsubsection{Segments across the Human–Robot Boundary}
\label{sec:3.3.4}

Passing the post-demonstration fast weights to the robot phase guarantees only continuity of memory values. For robot action losses to supervise demonstration writes, the computation graph must also span the phase boundary. Truncating at that boundary blocks gradients from robot losses to earlier human writes along the fast-weight path.

We therefore construct a transition segment at the boundary. It contains the final human anchors and the first robot anchors. No truncation occurs at the human–robot boundary within this segment. Outer supervision comes from robot anchors, but gradients can pass through consecutive fast-weight updates to human writes within the segment.

Earlier human anchors are processed chronologically as a memory-writing prefix. They may be segmented with gradient truncation, while fast-weight values are retained. At the end of the transition segment, we backpropagate and truncate the graph. The remaining robot trajectory uses regular TBPTT segments, whose gradients accumulate as in Figure~\ref{fig:tbptt} before the final slow-weight update.

Human and robot anchors in the transition segment retain their respective sampling frequencies. Their anchor counts and write frequencies are configured independently. Changing a write frequency therefore also changes the time span covered by gradients.

This arrangement lets robot losses in the transition segment supervise some human writes, but does not connect the entire human demonstration and robot trajectory through an unbroken gradient path. Earlier human information and later robot actions may still be separated by truncation boundaries. Memory values can persist, while the temporal reach of supervision remains limited by segmentation.

\begin{algorithm}[htbp]
\caption{TBPTT with Human Demonstrations and Robot Trajectories}
\label{alg:human-robot-tbptt}
\small
\noindent\textbf{Require:} Paired-trajectory loader \(D\), outer-loop parameters \(\Theta\)\par
\noindent\textbf{Require:} Learnable initial fast weights \(F_{\mathrm{init}}\in\Theta\), optimizer \(\mathrm{opt}\)\par
\noindent\textbf{Require:} Human sampling frequency \(f_H\), robot chunk length \(C\)\par
\noindent\textbf{Require:} Segment length \(S\), transition sizes \(K_H\) and \(K_R\)\par
\begin{enumerate}
\renewcommand{\labelenumi}{\arabic{enumi}:}
\setlength{\itemsep}{1pt}
\setlength{\parskip}{0pt}
\setlength{\parsep}{0pt}
\item \textbf{for} trajectory batch \(B\) in \(D\) \textbf{do}
\item \hspace*{1em}\(\mathrm{opt}.\operatorname{zero\_grad}()\)
\item \hspace*{1em}\textbf{for} \((\text{human demonstration},\text{robot trajectory})\) in \(B\) \textbf{do}
\item \hspace*{2em}\textbf{Step 1: Anchor sampling and segmentation}
\item \hspace*{2em}\(A_H \gets \operatorname{SampleHuman}(\text{human demonstration},f_H)\)
\item \hspace*{2em}\(A_R \gets \operatorname{SampleRobot}(\text{robot trajectory},C)\)
\item \hspace*{2em}\(P \gets \operatorname{BuildSegments}(A_H,A_R,S,K_H,K_R)\)
\item \hspace*{2em}\(F \gets \operatorname{CopyWithGradient}(F_{\mathrm{init}})\)
\item \hspace*{2em}\textbf{for} segment \(P_j\) in \(P\) \textbf{do}
\item \hspace*{3em}\textbf{Step 2: Memory propagation and task supervision}
\item \hspace*{3em}\(\mathrm{losses} \gets \varnothing\)
\item \hspace*{3em}\textbf{for} anchor \(a\) in \(P_j\) \textbf{do}
\item \hspace*{4em}\textbf{if} \(a\) is a human anchor \textbf{then}
\item \hspace*{5em}\(F \gets \operatorname{HumanWrite}(a,F;\Theta)\)
\item \hspace*{4em}\textbf{else}
\item \hspace*{5em}\(\varepsilon,\tau \gets \operatorname{SampleFlowMatchingNoiseAndTime}()\)
\item \hspace*{5em}\(\ell_a,F \gets \operatorname{RobotTrainStep}(a,F,\varepsilon,\tau;\Theta)\)
\item \hspace*{5em}\(\mathrm{losses} \gets \mathrm{losses}\cup\{(a,\ell_a)\}\)
\item \hspace*{4em}\textbf{end if}
\item \hspace*{3em}\textbf{end for}
\item \hspace*{3em}\textbf{Step 3: Backpropagation and gradient truncation}
\item \hspace*{3em}\textbf{if} \(\mathrm{losses}\ne\varnothing\) \textbf{then}
\item \hspace*{4em}\(L_j \gets \operatorname{WeightedTaskLoss}(\mathrm{losses},B)\)
\item \hspace*{4em}\(L_j.\operatorname{backward}()\)
\item \hspace*{3em}\textbf{end if}
\item \hspace*{3em}\(F \gets \operatorname{DetachAsLeaf}(F)\)
\item \hspace*{3em}\(\operatorname{ReleaseSegmentGraph}()\)
\item \hspace*{2em}\textbf{end for}
\item \hspace*{1em}\textbf{end for}
\item \hspace*{1em}\textbf{Step 4: Slow-weight update}
\item \hspace*{1em}\(\mathrm{opt}.\operatorname{step}()\)
\item \textbf{end for}
\end{enumerate}
\end{algorithm}



In Algorithm~\ref{alg:human-robot-tbptt}, BuildSegments constructs human-prefix segments, a transition segment, and subsequent robot segments in chronological order. The transition segment contains the last \(K_H\) human anchors and the first \(K_R\) robot anchors. Each anchor is processed once, and gradients are not truncated within the transition segment.

HumanWrite writes demonstration memory without computing a robot action supervision loss. RobotTrainStep runs a training forward pass for a robot anchor and returns its task loss and updated memory. Their internal TTT operations follow Algorithm~\ref{alg:ttt-layer}.

WeightedTaskLoss preserves the training objective's loss weights and normalization. DetachAsLeaf retains fast-weight values and truncates historical gradient paths while allowing subsequent inner updates to differentiate with respect to fast weights. Outer-loop gradients accumulate across segments, with one slow-weight update after the full trajectory batch.

\section{Experiments}
\label{sec:4}

This section follows the progression of our experiments. We first describe supervision, training objectives, and evaluation. We then present training on the original video demonstration dataset and the observation that changing demonstration memory did not change task decisions. To investigate this, we examine gradient coverage on a short-video demonstration dataset and try contrastive regularization. Finally, we reduce video demonstrations to single-frame cues to study learning with simpler conditions.

\subsection{Supervision, Objectives, and Evaluation}
\label{sec:4.1}

\subsubsection{GT-State and AR-State Open-Loop Evaluation}
\label{sec:4.1.1}

We evaluate action prediction using two open-loop protocols: GT-state and AR-state. Both read images along dataset trajectories and perform the full action-denoising procedure. They differ in whether the next prediction uses the ground-truth robot state or the prediction from the previous action chunk. For TTT models, both protocols carry fast weights along the trajectory and reset them at the start of a new trajectory.

\textbf{GT-state open-loop evaluation.} Each anchor uses the dataset image and ground-truth robot state to predict the next action chunk. Prediction errors from the previous chunk do not directly enter the next anchor's state input. This protocol primarily measures action prediction under demonstrated observations and states. In a memory model, memory formed at earlier anchors can still influence later predictions.

\textbf{AR-state open-loop evaluation.} Each trajectory starts from its ground-truth initial state. Thereafter, the final valid predicted action of the previous chunk approximates the next anchor's robot state. This adds a feedback path: the current action prediction affects the next state, which affects subsequent actions and memory updates. The protocol examines how state errors propagate across successive predictions.

AR-state images still come from the dataset and do not change with predicted actions. It is therefore open-loop evaluation with autoregressive states, not an equivalent of closed-loop execution in a real environment. The target state implied by a predicted action is also not necessarily physically reachable.

Both protocols recover absolute actions in their original physical units before computing mean absolute error (MAE) and root mean squared error (RMSE). Predicted action increments are decoded relative to the current input state, while action labels are always decoded relative to the dataset's ground-truth state. Padding and invalid trailing steps are excluded, and joint and gripper errors are reported separately. Both protocols use identical evaluation trajectories, sampled anchors, and denoising settings.

We use training loss to monitor optimization and compare models using physical action errors after full denoising.

\subsubsection{Misalignment with Cross-Step Memory}
\label{sec:4.1.2}

Current training uses dataset images and ground-truth robot states at every anchor, supervised by the corresponding demonstrated actions. In terms of inputs and supervision, this is a GT-state training objective. With TTT, fast weights carry along the trajectory and connect multiple anchors through memory. We must therefore also consider whether earlier memory is consistent with the state and observation at the next anchor.

\begin{figure}[htbp]
\centering
\includegraphics[width=0.95\linewidth,height=0.58\textheight,keepaspectratio]{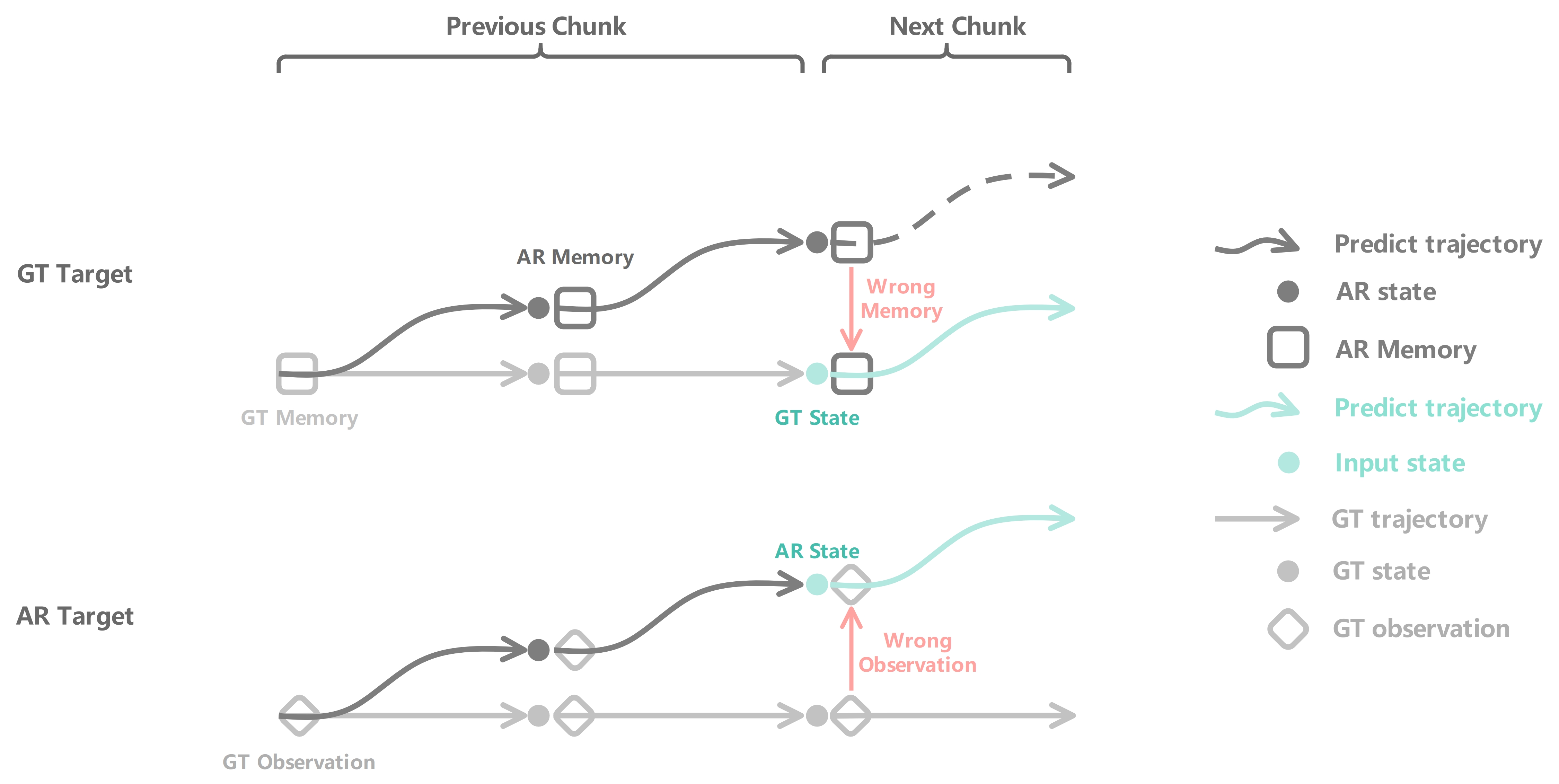}

\caption{\textbf{Misalignment under two state-input conditions.} In the upper GT-state case, earlier memory carries forward, but subsequent states still come from demonstrations rather than the execution of the model's decisions. In the lower AR-state case, predicted actions feed into subsequent states, while visual observations still come from demonstrations and may conflict with the propagated states. The predicted trajectories and memory illustrate these relationships, not actual environment interactions during training.}
\label{fig:gt-ar-misalignment}
\end{figure}

GT-state and AR-state involve different forms of misalignment, as shown in Figure~\ref{fig:gt-ar-misalignment}.

\textbf{GT-state: mismatch between historical memory and subsequent states.}

Under GT-state training, the next anchor receives the correct demonstration image and state even if the current prediction is wrong. Fast weights, however, are not reset alongside the state input and continue to carry earlier memory.

Internal representations involved in earlier predictions may therefore affect later computation, while state changes caused by executing those predictions are absent from training. Historical memory is paired with demonstration states during training and with actual execution states during deployment, potentially producing different input combinations.

Our implementation does not directly write final predicted actions back into fast weights. The mismatch should therefore not be interpreted simply as storing wrong actions in memory. The relevant issues are the persistence of internal representations and the effects of execution feedback on later observations, states, and memory formation.

GT-state training still teaches the model to predict demonstrated actions using evolving memory and correct observations and states. It does not directly train the model to continue the task from altered states and observations after executing its own erroneous actions.

\textbf{AR-state: mismatch between propagated states and visual observations.}

AR-state uses the previous action-chunk prediction as the next state, allowing the consequences of past decisions to enter later inputs. This adds the action-to-state feedback path absent under GT-state conditions.

In offline data, however, visual observations still come from demonstration trajectories. Once a predicted state deviates from the ground truth, the model may receive an image of the demonstrated position together with a state at the predicted position. Fast weights continue to update from this combination.

Autoregressive states therefore do not guarantee alignment among memory, states, and observations. They add prediction feedback but also introduce observation–state mismatch. Replacing ground-truth states with predicted states alone does not reproduce the full feedback process of real execution.

\textbf{Choice for subsequent experiments}

A theoretically complete supervision and evaluation setup requires memory, observations, and states to be aligned. Unfortunately, in the supervised fine-tuning (SFT) setting considered here, only two of the three can be aligned at a time.

On the warm-up dataset, we compare baseline delta without TTT against TTT delta and TTT abs. TTT delta is closer to a GT-state objective, while TTT abs is closer to an AR-state objective. We use this action-representation comparison to examine the objective difference indirectly; the interpretation is detailed in Appendix~\ref{app:a}. Results after 6,000 training steps are shown below:

\begin{table}[htbp]
\centering
\caption{Warm-up comparison after 6,000 training steps.}
\label{tab:warmup}
\small
\renewcommand{\arraystretch}{1.15}
\begin{tabular*}{\linewidth}{@{\extracolsep{\fill}}lccc@{}}
\toprule
\textbf{Model} & \textbf{GT-state MAE \(\downarrow\)} & \textbf{AR-state MAE \(\downarrow\)} & \textbf{Real-robot completion rate \(\uparrow\)} \\
\midrule
baseline delta & 0.023604 & 0.099308 & \textbf{90\%} \\
TTT delta & \textbf{0.022631} & 0.091122 & 73\% \\
TTT abs & 0.028153 & \textbf{0.082412} & 64\% \\
\bottomrule
\end{tabular*}
\end{table}

Compared with TTT delta, TTT abs has lower AR-state open-loop error but a lower real-robot completion rate. The improved AR metric therefore did not translate into better task performance. Subsequent training retains GT-state supervision and incremental action representations.

\subsection{Initial Training with Video Demonstrations}
\label{sec:4.2}

After choosing the supervision conditions, we train on the original video demonstration dataset. This stage uses a cup-selection task with paired human videos and robot trajectories. The demonstration identifies the target cup, and the robot trajectory contains the corresponding grasp. Training on these pairs is intended to teach the relation between demonstration conditions and target selection.

The model first reads the human demonstration into fast weights, then carries that memory into robot-trajectory processing. Training uses GT-state supervision, incremental actions, and the TBPTT procedure in Section~\ref{sec:3.3}. Robot supervision constrains the corresponding human writes through memory updates retained within each segment.

\subsubsection{Training Setup}
\label{sec:4.2.1}

Task training has two stages. The first uses a cup-grasping dataset in which the robot picks up a cup and puts it down. After TTT is added, robot trajectories teach basic grasping and placing skills to support later target selection from demonstrations.

The second stage trains the cup-selection task on the original video demonstration dataset. The dataset contains 654 complete paired samples and 556,302 frames. Each pair consists of a human video identifying the target cup and a robot trajectory recording its grasp. Starting from first-stage weights, the model reads the demonstration, updates memory, and uses that memory to predict robot actions. Robot supervision passes through the retained graph to constrain demonstration writes and teach their relation to target selection. This stage fine-tunes all parameters and creates a new optimizer and learning-rate scheduler.

\textbf{InternVLA-A1-TTT}

We add TTT to every layer of InternVLA-A1's\cite{3} generation and action experts and compare two initializations: one without additional TTT pretraining and one with 20,000 steps of TTT pretraining. Both then undergo 30,000 first-stage steps and 60,000 second-stage steps. The second-stage split is made at the level of complete pairs, with 621 training and 33 validation samples.

Human demonstrations write to generation-expert memory at 1.5 Hz. Robot anchors are sampled every 30 control steps and predict 30-step action chunks. TBPTT segments contain 16 anchors; the transition segment contains the last 8 human anchors and the first 8 robot anchors.

\textbf{Pi0.5-TTT}

For Pi0.5\cite{2}, we perform 60,000 steps of TTT pretraining on the same InternData\cite{18} and retain the exponential moving average (EMA) weights. We compare initialization from these weights with initialization from the original official weights. Both undergo 30,000 first-stage steps and 30,000 second-stage steps.

Human demonstrations write memory at 1.5 Hz. Robot anchors are sampled every 30 control steps and predict 30-step action chunks. Each TBPTT segment contains 4 anchors.

\subsubsection{Experimental Results}
\label{sec:4.2.2}

Both InternVLA-A1 and Pi0.5 showed a fixed directional preference in real-robot tests. For our cup-selection tasks, success rate is the fraction of trials in which the robot selects the cup specified by the demonstration or cue. Both models scored below the theoretical rate of uniform selection among three candidate cups (1/3).

We evaluate the models using Section~\ref{sec:4.1} and additionally test whether demonstration memory affects task decisions. Skipping demonstration-memory writes causes only small changes in action error. In real-robot tests, models tend to select similar fixed targets with or without demonstrations. Although demonstrations are training conditions, the expected dependence on them was not observed in these tests.

We first consider the coverage of training segments. Key information appears early in the human demonstration, while robot trajectories usually begin with shared lifting and reaching motions before target-specific actions. The two events are far apart in time. A short transition segment may not cover both the key demonstration information and the target-selection action. Memory values persist, but the relevant decision loss may not reach the write that encoded the key information.

Another possibility is that the original fitting objective does not sufficiently encourage the model to distinguish demonstration conditions. It trains predictions under correct pairings to match demonstrated actions without explicitly comparing predictions under different demonstrations. We therefore consider contrastive regularization that retains the original fitting objective and penalizes incorrect pairings whose losses are lower than those of correct pairings.

These possibilities concern gradient coverage and the training objective, respectively. We investigate them by shortening sequences so that one segment covers the full human–robot pair, then trying contrastive constraints in that setting.

\subsection{Full-Trajectory Training on Short Sequences}
\label{sec:4.3}

We construct a short-video demonstration dataset to reduce the temporal gap between key information and task decisions. The task still requires selecting the target cup from a human demonstration, but both human and robot sequences are shorter.

One segment contains all human and robot anchors in a pair. No segment boundary truncates the graph within that pair, allowing robot decision losses to propagate through fast-weight updates to earlier human writes. This change is intended to test whether limited gradient coverage was a major reason for the weak use of demonstration conditions.

\subsubsection{Training Setup}
\label{sec:4.3.1}

This stage uses 273 short-video demonstration pairs, split by complete sample into 264 training and 9 validation pairs.

We add TTT to the original backbone weights and train directly on the short-video dataset, without the two-stage grasping-then-conditioning procedure in Section~\ref{sec:4.2}. Training retains GT-state supervision and incremental actions. Robot anchors are sampled every 50 control steps and predict 50-step chunks. This reduces the number of anchors so that complete trajectories fit in GPU memory during training.

Each sample contains a human demonstration and its robot trajectory. The model processes both phases chronologically and carries fast weights across the boundary. Shorter sequences let one segment cover all anchors, allowing robot action losses to propagate through the retained graph to writes throughout the demonstration.

\textbf{InternVLA-A1-TTT}

The model starts from the original InternVLA-A1 backbone weights, with TTT in every generation- and action-expert layer. Human demonstrations write generation-expert memory at 1.5 Hz, and both experts continue to read and update memory during the robot phase. Segments contain up to 16 anchors, and the full human–robot sequence of each training sample fits within this limit. No separate transition segment with fixed human and robot anchor counts is needed, and the gradient path remains intact within each sample. We train for 60,000 steps under this setting.

\subsubsection{Experimental Results}
\label{sec:4.3.2}

We train under the settings above. Although one segment covers each complete pair, the experiment still does not yield the expected behavior. InternVLA-A1 reaches cup-selection success rates of 25\% and 13\% after 10,000 and 60,000 steps, respectively.

We then try contrastive regularization on the same short-video dataset. Training retains the fitting objective for correct pairs and uses alternative demonstration–trajectory pairings to encourage discrimination between demonstration conditions. We observe no clear improvement over training without the regularizer in this experiment. The objectives, implementation, and detailed results are given in Appendix~\ref{app:b}.

Shortening sequences and adding conditional constraints did not produce the expected behavior. This led us to consider the demands of understanding video demonstrations and simplify the conditioning input further. The next stage replaces a video with a single image cue to test whether a more direct task cue can influence robot decisions.

\subsection{Task Training with Single-Frame Cues}
\label{sec:4.4}

In Section~\ref{sec:4.3}, shorter sequences, full-trajectory gradient connectivity, and contrastive regularization did not produce the expected dependence on demonstrations. We then considered that video demonstrations require not only retaining information but also understanding relations between frames. TTT provides a mechanism for cross-step memory, but whether the current training teaches video understanding and use remains unverified.

We therefore simplify the condition to a single-frame cue. The image explicitly marks the correct cup's position, removing the need to track target changes across video frames. The robot must still select the target from the cue. We use this simpler setting to examine whether the model can write the cue into memory and learn to use it for subsequent actions.

\subsubsection{Training Setup}
\label{sec:4.4.1}

The single-frame cue dataset contains 210 paired samples. Each retains only the final human-demonstration frame as a cue, together with the full robot trajectory. Two original trajectories without human demonstrations were removed. Using seed 42, we split complete samples at approximately 95:5 into 200 training and 10 validation samples.

This stage follows the two-stage procedure in Section~\ref{sec:4.2}, loading the corresponding first-stage weights before training on single-frame cues. The model reads the cue, updates memory, and carries that memory into robot action prediction. Training retains GT-state supervision and incremental actions.

\textbf{InternVLA-A1-TTT}

We compare two initializations. The first uses weights after 30,000 first-stage grasping steps without extra TTT pretraining. The second uses weights after 20,000 TTT pretraining steps followed by 30,000 first-stage grasping steps. Both use the same first-stage checkpoints as in Section~\ref{sec:4.2} and continue for 60,000 steps on the single-frame cue dataset. Their configurations are identical apart from initialization.

TTT is inserted into every generation- and action-expert layer. The action expert uses state and register tokens for writing, with 16 register tokens. The cue image writes to both experts' memories. The cue sampling frequency is set to 1 Hz, but each sample contains only one frame and therefore yields one cue anchor.

Robot anchors are sampled every 30 control steps and predict 30-step action chunks. Trailing anchors with fewer than 30 steps are retained. Starting at the cue anchor, TBPTT divides the sequence into consecutive segments of at most 16 anchors. Gradients remain connected within segments; fast-weight values carry across segments while the graph is truncated. Model parameters are updated once after each complete trajectory batch.

Both groups fine-tune all parameters without contrastive regularization or image augmentation. Training uses two GPUs, each processing two trajectories in parallel, for a global batch of four trajectories. The optimizer and scheduler are recreated for this stage, with a learning rate of \(5 \times 10^{-5}\), 50 warm-up steps, and a decay horizon of 60,000 steps.

\textbf{Pi0.5-TTT}

We compare two models whose TTT layers use the constrained Fast MLP and multi-head core, respectively. They use the same first-stage checkpoints as in Section~\ref{sec:4.2} and each continue for 60,000 steps on the single-frame cue dataset.

Robot anchors are sampled every 30 control steps and predict 30-step action chunks. Trailing anchors with fewer than 30 steps are retained. Starting at the cue anchor, TBPTT divides the sequence into consecutive segments of at most 4 anchors. Gradients remain connected within segments; fast-weight values carry across segments while the graph is truncated. Model parameters are updated once after each complete trajectory batch.

Both groups fine-tune all parameters. Training uses two GPUs, each processing two trajectories in parallel, for a global batch of four trajectories. The optimizer and scheduler are recreated for this stage, with a learning rate of \(2 \times 10^{-5}\), 2,000 warm-up steps, and a decay horizon of 60,000 steps.

\subsubsection{Experimental Results}
\label{sec:4.4.2}

In real-robot tests with single-frame cues, neither InternVLA-A1-TTT nor Pi0.5-TTT showed the expected dependence on the cue. With cup positions fixed, both models tended to grasp a cup at a fixed position rather than consistently change their selection with the marked target. Both models had cup-selection success rates below 33\%.

These results indicate that stable dependence on single-frame cues has not yet emerged, but do not establish that TTT cannot remember single-frame information. Reducing a video to one frame simplifies the input but does not guarantee that target information is sufficiently written, retained, and used for action generation.

One possibility is that a single cue write is insufficient to form useful target memory. Alternatively, the information may be written but weakened by later robot-phase updates or not effectively read by the action branch. If the target-selection action falls in a later TBPTT segment, its supervision also cannot propagate directly back to the cue write. Future checks could repeat the same cue write, keep memory read-only during the robot phase, and assess target decodability and decision effects as in Section~\ref{sec:6.3} to distinguish these possibilities.

\section{Conclusion}
\label{sec:5}

Our main contribution is to reproduce and clearly describe the TTT architecture and its training procedure for combining human demonstrations with robot trajectories in VLA models. Training infrastructure constraints have limited the scale of pretraining and training on memory-task data, leaving TTT's memory capabilities insufficiently trained and evaluated.

We integrate TTT into InternVLA-A1 and Pi0.5 and progress from video demonstrations to short videos and single-frame cues. In the tests performed, models respond weakly to demonstration conditions and show fixed-position cup preferences in real-robot execution. After changes to sequence length, gradient coverage, and training constraints, the evidence still does not establish that the models learned to use demonstration memory for target selection. The objective comparison also shows that lower open-loop action errors do not necessarily correspond to better real-robot task performance.

We cannot yet determine whether the main issue lies in writing information, retaining it, or using it for action generation, nor have we verified whether scaling up training would improve these observations. We document the implementation, configurations, evaluation procedures, and unresolved observations as a record for further inspection and replication.

\section{Discussion}
\label{sec:6}

\subsection{TTT Layer Design Choices}
\label{sec:6.1}

We use a two-layer GELU MLP for fast-weight memory, with a hidden dimension four times the input dimension. This follows the original TTT-MLP\cite{1} configuration; RoboTTT\cite{4} also uses a two-layer GELU MLP. It provides nonlinear expressiveness while keeping the inner update path shallow. We use it as a starting point and have not established its relative advantage on our robot tasks through systematic ablations.

ViT\textsuperscript{3}\cite{11} provides guidance on other visual TTT configurations. Its results suggest that increasing inner-model width can increase capacity, while greater depth may make optimization harder. It uses a single full-batch inner update with a fixed learning rate of 1. Our implementation likewise aggregates write tokens within a call, takes one fast-weight update, and fixes the inner learning rate at 1. Fast weights still pass sequentially across temporal anchors.

These observations can guide configuration choices but do not directly establish their suitability for our robot tasks. The relevant MLP experiments in ViT\textsuperscript{3} use SiLU, and its final architecture also introduces gating and convolutions. It therefore does not establish a two-layer GELU MLP as the best inner model. Robot state, action, and visual representations also have different distributions, and fast weights must update across time. The effects of learning rate, inner-model capacity, and update count on long-term memory require comparisons under matched tasks and training conditions.

\subsection{TTT Placement and Layer Coverage}
\label{sec:6.2}

We add TTT to every layer of InternVLA-A1's generation and action experts, leaving the understanding expert unchanged. TTT follows the joint-attention output projection and first residual connection and precedes the original feedforward sublayer. It therefore receives attention-fused features and passes memory readouts into subsequent computation. The generation and action branches maintain separate fast weights, allowing memory to participate in both visual and action generation.

This provides two paths for memory use, but their coexistence does not establish an effective division of labor. Generation-expert memory may affect later actions through visual representations, while action-expert memory directly participates in processing action representations. Whether the branches use different information, duplicate it, or interfere with each other requires branch ablations and demonstration-condition controls.

The number of TTT layers should be considered together with their placement. Each added layer introduces independent fast weights and additional inner updates and outer-loop backpropagation. More layers provide more memory interfaces but increase computation and training GPU memory; evidence that these costs lead to better memory use remains limited. Future comparisons could vary layer coverage under fixed data, training procedures, and fast-weight configurations, testing generation-only, action-only, and joint integration.

\subsection{Memory Evaluation}
\label{sec:6.3}

For interpretability, action error, memory decodability, and memory's effect on decisions can be evaluated separately. For suitable tasks, a linear probe can predict labels such as grasp direction on independent trajectories to test whether target information is retained across layers and time. A fixed query reduces the influence of the current observation, while features before and after gating under actual queries reveal changes in readout. Probe accuracy measures decodability, not memory capacity or decision contribution directly.

Comparisons across histories hold the current image, state, prompt, and noise fixed while substituting correct-history memory, incorrect-history memory, or initial fast weights. Comparing read-only history with continued writes from the current observation can test whether current writes overwrite history. The main outcomes are target selection and any improvement from correct history; history swaps should match posture, layout, and task phase. Fast-weight distances, drift, and gate magnitudes are auxiliary diagnostics, not standalone evidence of effective memory.

However, the widespread use of randomized object positions in our later datasets makes linear-probe interpretability analysis difficult.

\subsection{Training Scale and Limitations}
\label{sec:6.4}

Adding TTT gives a single-step model the structure to store and read information across steps, but using history must still be learned. The model must learn what to write, how to retain it through later updates, and when to use memory for action generation. Extending a single-step model into one that uses memory effectively therefore introduces new learning requirements, whose data scale and training procedures remain to be examined.

InternVLA-A1\cite{3} is pretrained on approximately 692 million frames spanning simulated manipulation, real robot trajectories, and human videos. In comparison, each of our three datasets for learning demonstration-conditioned behavior contains a few hundred pairs, as shown below.

\begin{table}[htbp]
\centering
\caption{Backbone pretraining and demonstration-conditioned task data.}
\label{tab:data-scale}
\small
\renewcommand{\arraystretch}{1.15}
\begin{tabular*}{\linewidth}{@{\extracolsep{\fill}}lccc@{}}
\toprule
\textbf{Data setting} & \textbf{Dataset size} \\
\midrule
InternVLA-A1 backbone pretraining & Approximately 692 million frames \\
Original video demonstration task & 654 pairs, 556,302 frames; 621 training pairs \\
Short-video demonstration task & 273 pairs; 264 training pairs \\
Single-frame cue task & 210 pairs; 200 training pairs \\
\bottomrule
\end{tabular*}
\end{table}

We train the added memory mechanism on existing backbone weights, so these task datasets do not represent all data previously seen by the model. Compared with the diverse manipulation covered by backbone pretraining, our demonstration-conditioned training centers on cup selection and covers a limited range of historical dependencies. Broader sequence data and larger-scale training may help the model learn to use memory, but the required scale cannot yet be determined from these experiments. Our observations therefore apply primarily to the current tasks and training settings.

\section*{Acknowledgments}
We acknowledge the support provided by Standard Robots.

\section*{Author Contributions}

Qiangong Zhou, Jiafeng Liang, and Yimin Dong were the primary contributors and contributed equally to this work.

\textbf{Qiangong Zhou}: Architecture design, code development, and TTT experiments on InternVLA-A1.
\textbf{Jiafeng Liang}: Architecture design, code development, and TTT experiments on Pi0.5.
\textbf{Yimin Dong}: Architecture design and experimental discussions.
\textbf{Yang He}: Infrastructure support for early experiments.
\textbf{Hongxiang Li}: Project supervision.

\clearpage
\bibliographystyle{unsrt}  
\bibliography{template}

\clearpage
\appendix
\section*{Appendix}

\section{Action Representations and GT/AR Objectives}
\label{app:a}

The main text compares TTT delta and TTT abs to examine GT-state and AR-state objectives indirectly. Here, “closer to” refers to how the action target depends on the current reference state and which direction it specifies after the state deviates from the demonstration. Both groups still train with ground-truth images and states; only the action representation differs. TTT abs is therefore not equivalent to AR-state training that directly uses predicted states.

TTT delta learns action increments relative to the current ground-truth state. Under GT-state conditions, training and decoding share the same reference state, so increments recover the demonstrated actions. Under AR-state conditions, decoding instead uses the state predicted by the previous action chunk. If that state already has an error and the model outputs the original increment, the error may persist into later actions. We regard this dependence on the ground-truth reference state as closer to a GT-state objective.

\begin{figure}[htbp]
\centering
\includegraphics[width=0.75\linewidth,height=0.58\textheight,keepaspectratio]{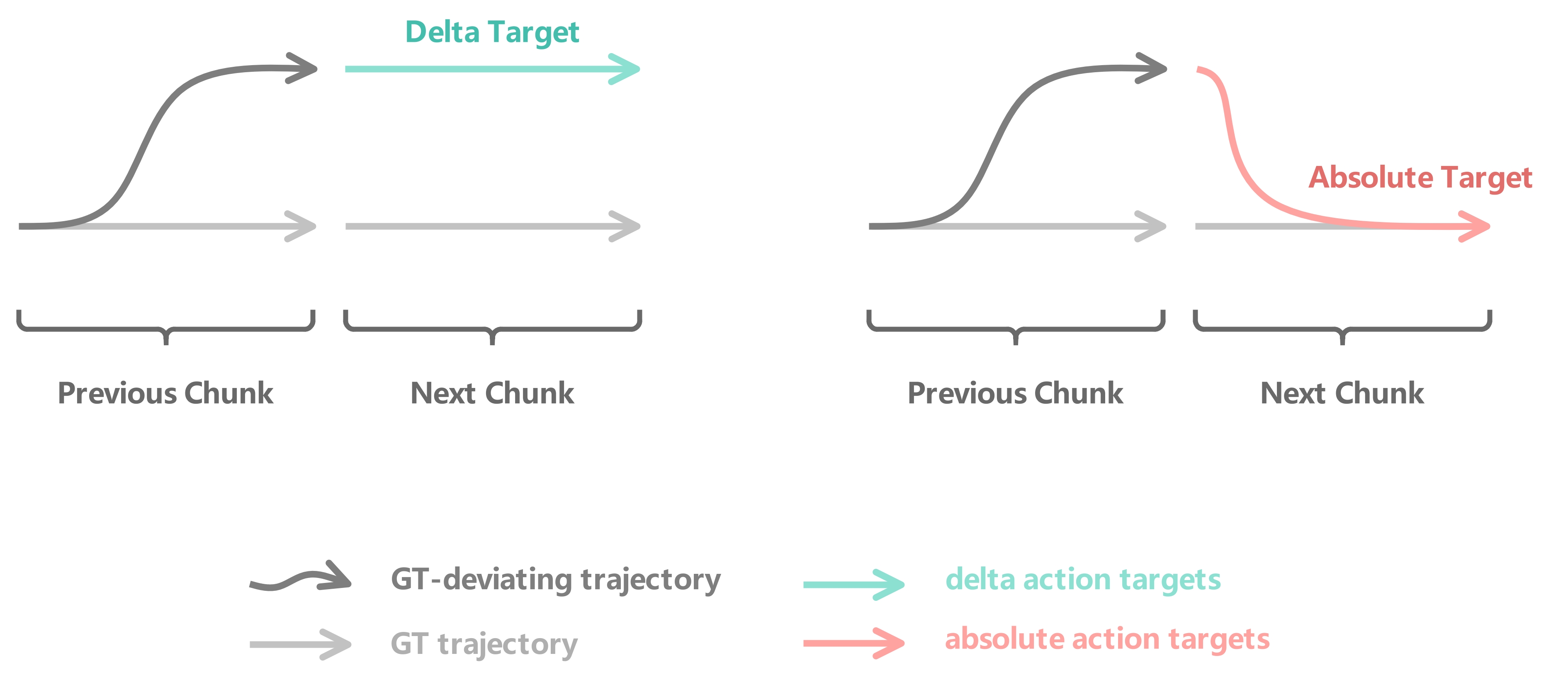}
\caption{\textbf{Incremental and absolute action targets during state autoregression.} The gray straight line denotes the demonstration trajectory, and the dark gray curve denotes deviation caused by the previous action chunk. On the left, green arrows apply the original increments to the deviated state, potentially preserving the existing position error. On the right, red arrows keep absolute targets at the demonstrated positions. The figure illustrates target geometry, not learned corrective behavior during real execution.}
\label{fig:action-targets}
\end{figure}

TTT abs directly learns absolute action targets. Even if the propagated state deviates from the demonstration, the target remains the original absolute position, without adding the current state during decoding. In this sense of continuing to target the demonstrated position from a deviated state, we regard it as an attempt closer to an AR-state objective. Figure~\ref{fig:action-targets} illustrates the difference after state deviation.

For example, suppose the next anchor's ground-truth joint position is 1.0, the propagated state is 0.7, and subsequent demonstrated actions remain at 1.0. The two action representations then behave as follows:

\begin{table}[htbp]
\centering
\caption{Action representations after state deviation.}
\label{tab:action-representations}
\small
\renewcommand{\arraystretch}{1.2}
\begin{tabular}{@{}
    >{\raggedright\arraybackslash}
    p{\dimexpr0.20\linewidth-2\tabcolsep\relax}
    >{\centering\arraybackslash}
    p{\dimexpr0.35\linewidth-2\tabcolsep\relax}
    >{\centering\arraybackslash}
    p{0.45\linewidth}
@{}}
\toprule
\textbf{Action representation}
&
\textbf{Training target relative to the ground-truth state}
&
\textbf{Absolute action if the same value is output after deviation}
\\
\midrule
delta & 0.0 & \(0.7 + 0.0 = 0.7\) \\
abs   & 1.0 & 1.0 \\
\bottomrule
\end{tabular}
\end{table}

With delta, reaching the same absolute target of 1.0 from 0.7 requires an increment of 0.3 rather than the zero increment in the training sample. With abs, the target remains 1.0. This motivates our expectation that absolute actions could reduce error propagation in AR-state open-loop evaluation.

Changing action representation does not supply actual observations at deviated states or direct supervision for recovery behavior. In the main experiment, TTT abs has lower AR-state MAE than TTT delta but a lower real-robot completion rate. This comparison therefore concerns how action representations affect target dependence and error propagation; it does not establish that AR-state training itself is better or worse than GT-state training.

\section{Contrastive Regularization: Design, Implementation, and Results}
\label{app:b}

The original trajectory-fitting objective predicts corresponding actions under the correct demonstration but does not explicitly compare demonstration conditions. We try trajectory contrastive regularization and memory separation regularization to strengthen this constraint. The former compares action prediction errors under different demonstrations; the latter encourages distinct fast-weight representations across demonstrations. Both are additional constraints, and the original correct-pair training objective is retained.

\subsection{Objectives and Sample Pairing}
\label{sec:b.1}

Let a global batch contain \(B\) valid human–robot pairs, denoted by \((H_i,R_i)\). All samples use the same representation and pairing rules, regardless of whether they reside on the same GPU. Define \(L_{ij}\) as the flow-matching loss against the ground-truth actions of \(R_i\), when predicting \(R_i\) using memory formed from \(H_j\). \(L_{ii}\) denotes a correct pairing, and \(L_{ij}\) with \(i\ne j\) denotes a cross-pairing within the batch.

For different pairings of the same robot trajectory, we hold robot images, language instructions, states, ground-truth actions, and flow-matching noise and timesteps fixed, replacing only demonstration memory. For each trajectory, losses are computed over valid actions and then averaged over valid anchors. Trajectories contribute equally within the batch.

\textbf{Trajectory contrastive regularization.} The fitting objective for correct pairs is:

\begin{equation}
\label{eq:fit}
\mathcal L_{\mathrm{fit}}
=
\frac{1}{B}\sum_i L_{ii}.
\end{equation}

We additionally compare losses for the same robot trajectory under correct and incorrect demonstrations:

\begin{equation}
\label{eq:contrastive}
\mathcal L_{\mathrm{contrast}}
=
\frac{1}{B(B-1)}
\sum_i\sum_{j\ne i}
\max(0,L_{ii}-L_{ij}).
\end{equation}

The term penalizes an incorrect demonstration when its loss is lower than that of the correct demonstration; otherwise, the penalty is zero. This implementation uses no additional margin and also has zero gradient when the losses are equal.

The comparison holds the robot trajectory fixed. For example, \(L_{21}\), obtained by predicting \(R_2\) from \(H_1\), is compared with the correct-pair loss \(L_{22}\). The constraint asks whether the correct demonstration is more helpful for the same robot prediction problem.

\textbf{Memory separation regularization.} Let \(F_i^H\) be the fast weights after reading the human demonstration and \(F_i^0\) their initial values. We extract and normalize the change caused by the demonstration:

\begin{equation}
\label{eq:memory-features}
z_i=
\frac{\operatorname{vec}(F_i^H-F_i^0)}
{\max\!\left(
\|\operatorname{vec}(F_i^H-F_i^0)\|_2,\epsilon
\right)}.
\end{equation}

The features include only TTT branches that the human demonstration actually writes to. Fast-weight changes from enabled layers are concatenated before normalization. The memory regularizer between demonstrations is:

\begin{equation}
\label{eq:memory-regularizer}
\mathcal L_{\mathrm{memory}}
=
\frac{2}{B(B-1)}
\sum_{i<j}
\exp\left(
-\frac{\|z_i-z_j\|_2^2}{\tau}
\right).
\end{equation}

More similar memory representations incur larger penalties. Normalization prevents reducing the penalty solely by increasing fast-weight magnitudes. The objective encourages distinct representations but does not by itself ensure that robot actions use those differences.

Training also retains the original auxiliary visual-generation loss:

\begin{equation}
\label{eq:base-loss}
\mathcal L_{\mathrm{base}}
=
\mathcal L_{\mathrm{fit}}
+
\lambda_{\mathrm{gen}}\mathcal L_{\mathrm{gen}},
\end{equation}

\begin{equation}
\label{eq:total-loss}
\mathcal L_{\mathrm{total}}
=
\mathcal L_{\mathrm{base}}
+
\lambda_{\mathrm{contrast}}\mathcal L_{\mathrm{contrast}}
+
\lambda_{\mathrm{memory}}\mathcal L_{\mathrm{memory}}.
\end{equation}

Trajectory contrastive scores use only action losses, excluding visual-generation losses. Gradient routing for each term is controlled separately, as described next.

\subsection{Computation and Gradient Flow}
\label{sec:b.2}

\begin{figure}[htbp]
\centering
\includegraphics[width=0.75\linewidth,height=0.58\textheight,keepaspectratio]{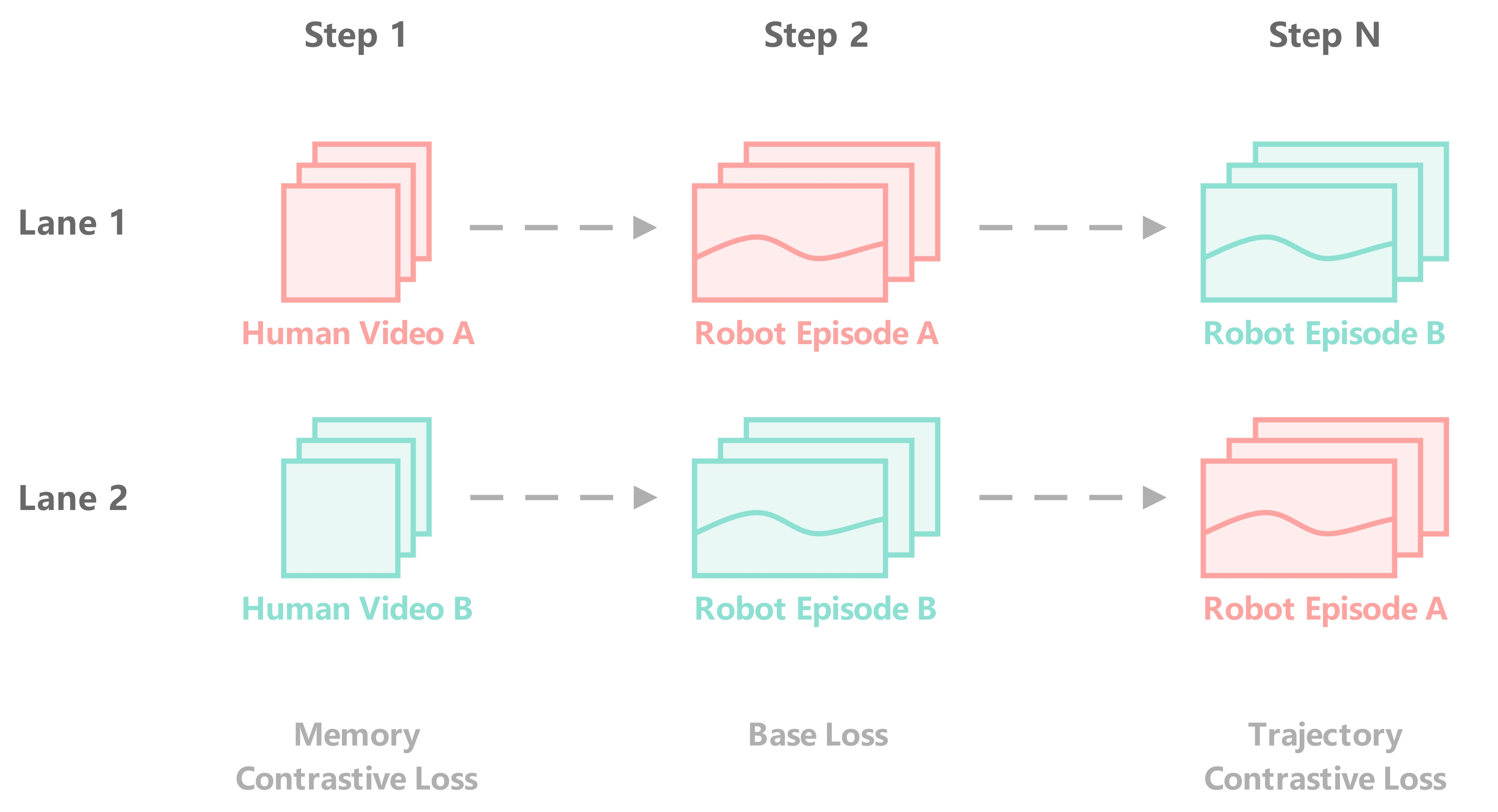}
\caption{\textbf{Computation of the contrastive regularizers.} First, each human demonstration is read, its terminal fast weights and graph are retained, and the memory separation regularizer is computed. Second, correct robot pairings are processed to compute the base training loss. Cross-pairings are then evaluated for the trajectory contrastive regularizer. Horizontal arrows indicate computation order; each robot branch restarts from the corresponding demonstration's terminal memory rather than inheriting updates from the preceding robot branch.}
\label{fig:contrastive}
\end{figure}

Full cross-pairing evaluates \(B^2\) human–robot combinations. To control GPU memory use, we retain human-demonstration graphs and process robot branches one at a time. Figure~\ref{fig:contrastive} shows the procedure.

Each human demonstration first performs one forward write pass, with its full computation graph retained. We then process correct pairings and save their action losses and gradients with respect to the demonstration's terminal fast weights. The base loss trains the robot prediction path and propagates gradients to human writes through the fast-weight interface.

For cross-pairings, each robot branch starts from an independent copy of the corresponding demonstration's terminal memory. Once its loss and gradients at the memory input are computed, the robot graph can be released. Gradients from different branches accumulate at the terminal human fast weights and are then propagated together through the retained human graph.

Both the positive and negative sides of the trajectory contrastive term are differentiated, but its gradients return to human writes only through demonstration memory. They do not directly update parameters in the robot prediction branch. The memory separation regularizer backpropagates through both human branches. The total objective therefore uses explicit gradient routing rather than an unrestricted gradient of the summed loss over all shared parameters.

This scheme retains the human graphs, the current robot graph, and the corresponding gradient buffers at the memory input, without storing every cross-pair robot graph at once. Parameters are updated once after all gradient contributions are accumulated. It reduces the number of simultaneously retained graphs, not the total number of pair evaluations.

Gradient paths must remain intact within both human and robot sequences. If the robot trajectory is truncated, later action losses cannot cross the truncation point to reach demonstration writes, even if the human graph is retained. Releasing graphs branch by branch and truncating gradients through time are different operations.

\subsection{Experimental Setup and Results}
\label{sec:b.3}

\textbf{Experimental setup}

We try contrastive regularization on the short-video demonstration task in Section~\ref{sec:4.3}. Unlike the original task, it removes the prolonged sequence of repeated cup-position swaps. The human demonstration contains only one cup-covering action, and the robot must select the target cup from it. This shortens the gap between demonstration information and robot target selection and reduces the demonstration's complexity.

Both groups train directly from the original backbone weights on the demonstration-conditioned task, without first-stage grasping training. We compare checkpoints after 10,000 steps of standard training and training with contrastive regularization.

The experiments use InternVLA-A1 with TTT in every generation- and action-expert layer. Human demonstrations write generation-expert memory at 1.5 Hz. Robot anchors are sampled every 50 control steps and predict 50-step action chunks. Standard training uses 16-anchor segments, each large enough to contain the selected sample's full human–robot sequence. Regularized training explicitly retains the full human–robot graph so that regularizer gradients from robot trajectories can reach demonstration writes.

Both groups fine-tune all parameters using incremental actions and a GT-state objective. The main settings are listed below.

\begin{table}[htbp]
\centering
\caption{Settings for the contrastive regularization comparison.}
\label{tab:contrastive-settings}
\small
\renewcommand{\arraystretch}{1.15}
\begin{tabular*}{\linewidth}{@{\extracolsep{\fill}}lccc@{}}
\toprule
\textbf{Setting} & \textbf{Value} \\
\midrule
Learning rate & \(5 \times 10^{-5}\) \\
Learning-rate warm-up steps & 50 \\
Weight decay & 0.01 \\
Parallel trajectories per GPU & 2 \\
Generation loss weight & 0.01 \\
Fast-weight computation precision & FP32 \\
Trajectory contrastive regularizer weight & 1,000, regularized group only \\
Memory separation regularizer weight & 0.1, regularized group only \\
Memory separation regularizer temperature & 1.0, regularized group only \\
\bottomrule
\end{tabular*}
\end{table}

The two regularizer weights bring the weighted losses to comparable orders of magnitude. The regularized group retains the original action-fitting and generation losses and adds the trajectory contrastive and memory separation terms. The trajectory term is a zero-margin hinge that penalizes only incorrect pairings with lower losses than correct pairings. Standard training aggregates action losses over valid robot anchors, while regularized training first averages within each trajectory and then across trajectories. The groups therefore also differ in loss aggregation.

With contrastive regularization, training runs at approximately 20\% of the standard training speed. For the same number of steps, elapsed time is approximately five times longer.

\textbf{Open-loop evaluation}

The table reports MAE for left-arm joint actions in absolute coordinates, measured in radians; lower values are better. GT-state and AR-state evaluation follow Section~\ref{sec:4.1}.

\begin{table}[htbp]
\centering
\caption{Open-loop MAE after 10,000 training steps.}
\label{tab:contrastive-open-loop}
\small
\renewcommand{\arraystretch}{1.15}
\begin{tabular*}{\linewidth}{@{\extracolsep{\fill}}>{\raggedright\arraybackslash}p{0.28\linewidth}cccc@{}}
\toprule
\textbf{Training setup} & \textbf{\shortstack{Training\\GT-state MAE \(\downarrow\)}} & \textbf{\shortstack{Training\\AR-state MAE \(\downarrow\)}} & \textbf{\shortstack{Validation\\GT-state MAE \(\downarrow\)}} & \textbf{\shortstack{Validation\\AR-state MAE \(\downarrow\)}} \\
\midrule
Standard training (10,000 steps) & 0.007650 & 0.024754 & 0.041284 & 0.107719 \\
Contrastive regularization (10,000 steps) & 0.008720 & 0.039244 & 0.044833 & 0.118272 \\
\bottomrule
\end{tabular*}
\end{table}

With contrastive regularization, both open-loop errors are higher on the training and validation sets than under standard training.

\textbf{With and without human demonstrations}

To test memory use, we compare reading demonstrations with skipping demonstration-memory writes on the validation set. Demonstration benefit is defined as MAE without a demonstration minus MAE with a demonstration, divided by MAE without a demonstration. A positive value indicates that adding the demonstration reduces error.

\begin{table}[htbp]
\centering
\caption{Validation MAE with and without demonstrations after 10,000 training steps.}
\label{tab:demonstration-ablation}
\small
\setlength{\tabcolsep}{3pt}
\renewcommand{\arraystretch}{1.15}
\begin{tabular*}{\linewidth}{@{\extracolsep{\fill}}>{\raggedright\arraybackslash}p{0.22\linewidth}cccccc@{}}
\toprule
\textbf{Training setup} & \multicolumn{3}{c}{\textbf{GT-state}} & \multicolumn{3}{c}{\textbf{AR-state}} \\
\cmidrule(lr){2-4}\cmidrule(lr){5-7}
& \textbf{\shortstack{With\\demonstration}} & \textbf{\shortstack{Without\\demonstration}} & \textbf{\shortstack{Demonstration\\benefit}} & \textbf{\shortstack{With\\demonstration}} & \textbf{\shortstack{Without\\demonstration}} & \textbf{\shortstack{Demonstration\\benefit}} \\
\midrule
Standard training (10,000 steps) & 0.041284 & 0.041287 & +0.0064\% & 0.107719 & 0.107573 & \ensuremath{-}0.1362\% \\
Contrastive regularization (10,000 steps) & 0.044833 & 0.044785 & \ensuremath{-}0.1072\% & 0.118272 & 0.118382 & +0.0930\% \\
\bottomrule
\end{tabular*}
\end{table}

For both checkpoints, the relative error changes with and without demonstrations are below 0.15\% in magnitude and have inconsistent signs. Contrastive regularization also does not produce a consistent demonstration benefit. These open-loop results do not provide evidence that the models effectively use demonstration memory.

\textbf{Real-robot results and behavior}

Real-robot evaluation uses the three-cup task with human demonstration videos. A trial succeeds when the robot selects the cup specified by the demonstration, with results shown below.

\begin{table}[htbp]
\centering
\caption{Real-robot cup-selection success after 10,000 training steps.}
\label{tab:contrastive-real-robot}
\small
\renewcommand{\arraystretch}{1.15}
\begin{tabular*}{\linewidth}{@{\extracolsep{\fill}}lccc@{}}
\toprule
\textbf{Training setup} & \textbf{Cup-selection success rate \(\uparrow\)} \\
\midrule
Standard training (10,000 steps) & 25\% \\
Contrastive regularization (10,000 steps) & 12.5\% \\
\bottomrule
\end{tabular*}
\end{table}

The regularized checkpoint at 10,000 steps does not outperform standard training at the same step. It also exhibits repeated movements and hesitation during grasping on the real robot. The available records do not establish the cause of this behavior.

Under this setting, contrastive regularization increases training cost without improving open-loop error, demonstration benefit, or real-robot cup-selection success. This observation is limited to the task and configuration tested and does not determine performance under other regularizer weights or training settings.

\end{document}